\documentclass[journal]{IEEEtran}
\usepackage{lineno,hyperref}

\usepackage{epsfig}
\usepackage{amsmath}
\usepackage{multirow}
\usepackage{threeparttable}
\usepackage[ruled,vlined]{algorithm2e}
\usepackage{amssymb}
\usepackage{amsthm}
\usepackage{balance}

\begin{document}

\title{Generic Semi-Supervised Adversarial Subject Translation for Sensor-Based Human Activity Recognition}

\author{Elnaz~Soleimani,        
        Ghazaleh~Khodabandelou,
        Abdelghani~Chibani,
        and~Yacine~Amirat
\thanks{E. Soleimani, G. Khodabandelou, A. Chibani, and Y. Amirat are with the Laboratoire Images, Signaux et Systmes Intelligents (LISSI), University Paris-Est Créteil, France. e-mail: elnaz.soleimani@u-pec.fr,ghazaleh.khodabandelou@u-pec.fr, chibani@u-pec.fr, amirat@u-pec.fr.}
}


\maketitle

\begin{abstract}
\boldmath
The performance of Human Activity Recognition (HAR) models, particularly deep neural networks, is highly contingent upon the availability of the massive amount of annotated training data which should be sufficiently labeled.
Though, data acquisition and manual annotation in the HAR domain are prohibitively expensive due to skilled human resource requirements in both steps. Hence, domain adaptation techniques have been proposed to adapt the knowledge from the existing source of data.
More recently, adversarial transfer learning methods have shown very promising results in image classification, yet limited for sensor-based HAR problems, which are still prone to the unfavorable effects of the imbalanced distribution of samples.
This paper presents a novel generic and robust approach for semi-supervised domain adaptation in HAR, which capitalizes on the advantages of the adversarial framework to tackle the shortcomings, by leveraging knowledge from annotated samples exclusively from the source subject and unlabeled ones of the target subject.
Extensive subject translation experiments are conducted on three large, middle, and small-size datasets with different levels of imbalance to assess the robustness and effectiveness of the proposed model to the scale as well as imbalance in the data.
The results demonstrate the effectiveness of our proposed algorithms over state-of-the-art methods, which led in up to 13\%, 4\%, and 13\% improvement of our high-level activities recognition metrics for Opportunity, LISSI, and PAMAP2 datasets, respectively. The LISSI dataset is the most challenging one owing to its less populated and imbalanced distribution.
Compared to the SA-GAN adversarial domain adaptation method, the proposed approach enhances the final classification performance with an average of 7.5\% for the three datasets, which emphasizes the effectiveness of micro-mini-batch training.  The manuscript provides a comprehensive evaluation of model performance, the explanation of the training procedure, the impact of sample population on the classifier performance, and the depiction of elements of the adversarial game.

\end{abstract}

\begin{IEEEkeywords}
Sensor Based Human Activity Recognition, Transfer Learning, Adversarial Domain Adaptation, Generative Adversarial Networks, Semi-Supervised Learning, Cross-Subject Transfer Learning
\end{IEEEkeywords}

\IEEEpeerreviewmaketitle

\section{Introduction}
\label{intro}
\IEEEPARstart{T}{hese} latest years, there is an unprecedented surge in integrating together Artificial Intelligence (AI) and Internet of Things (IoT) technologies in particular for ambient assisted living (AAL), Healthcare and ubiquitous robotics application domains \cite{FISCHER2020103285,wang2018deep,NUNEZ201880,chibani2013}. The considerable progress witnessed recently in AI  has been made, in various areas such as image and speech recognition, natural language processing and computer vision, thanks to the contributions of the machine learning community and the rapid growth and availability of high performance computing resources. The latter are increasingly stimulating data acquisition as well as the development of data stores and marketplaces in the web, which provide public and private datasets. Compared with traditional machine learning approaches, the unparalleled performance of deep learning methods is made possible thanks to their capability of learning from a large amount of data samples \cite{bengio2013deep}.

Nevertheless, there exceptionally exist some fields of study, which face limitations on data acquisition and labeling due to practical constraints. Therefore, it is not feasible to provide enough labeled data in many cases. Taking for instance the AAL domain and the problem of sensors based Human Activity Recognition (HAR) in particular, the data acquisition and labeling tasks require the implication of human annotators and the use of pervasive and intrusive sensors such as video cameras, which are making it more challenging to preserve the privacy of human subjects. Furthermore, the cost of the manual annotation is prohibitively expensive, especially for large-scale datasets.
Most of the machine learning models work based on the primary condition that samples of training (source domain) and test (target domain) set must be drawn from the same distribution and feature space. However, in many real-world applications such as HAR, this assumption cannot be held and consequently causes a dramatic decrease of models performance \cite{pan2010survey}. In this context, how a model that is trained on an initial amount of labeled data in a source domain can be adapted to generalize on unlabeled data in a target domain Today, it is considered that unlabelled data can give an indication of how a source domain and a target domain differ from each other. This information can be used by a classifier to modify its decisions in order to better generalize to the target domain \cite{Kouw2019}. Therefore, employing domain adaptation techniques could be beneficial for HAR in order to prevent models from suffering performance degrading when they are applied on new subjects or datasets.

Domain adaptation techniques have been considered as a way for automatic knowledge transfer from one domain to another to avoid the significant reduction of performance metrics, requiring as less amount of explicit training data as possible, in the target domain \cite{khan2018scaling}.
It has been employed in almost every deep learning model when the target dataset does not contain enough labeled data \cite{yosinski2014transferable}.
Based on the availability of annotated data in the source or target domain, adaptation methods are categorized into supervised (inductive), semi-supervised (transductive), and unsupervised ones.

This paper develops a transductive method for HAR, which is based on the initial SA-GAN adversarial approach proposed in \cite{Soleimani2019Cross}. The latter offers good results only in the case of a dataset that is well-balanced and contains a huge number of labeled samples. The present study, which is undertaken in the context of the European project Medolution \footnote{https://itea3.org/project/medolution.html}, has significantly enhanced the SA-GAN approach making it more generic and robust to prevent declining in performance, when there are not enough labeled data and the imbalanced class distribution of the available samples imposes additional obstacles. The proposed approach transfers knowledge from the source subject with distribution $P_{s}$ to the target subject distributed with $P_{t}$ using an adversarial framework. The new proposed model uses annotated data exclusively from the source domain and unlabeled data from target domain. The present study focus on the exploitation of inertial sensors' data, which are not intrusive and cope with privacy requirements.
The main contributions of the present attempt lie in:
\begin{itemize}

   \item Generic semi-supervised Adversarial Domain Adaptation technique that cope with cross-subject transfer learning problems in Human Activity Recognition domain. The proposed model benefits from convolutional neural networks to perform more generalized automatic feature extraction which is advantageous for classifying high-dimensional data of high-level activities. The proposed model is proven to be robust to the imbalance learning challenges by exploiting a micro-mini-batch learning strategy described in section \ref{adv_adap}.

  \item The proposed method have been extensively tested and evaluated by using cross-subject domain adaptation scenarios on three representative datasets; the two benchmark datasets, Opportunity \cite{chavarriaga2013opportunity} and PAMAP2 \cite{reiss2012introducing}, and a dataset on rehabilitation self exercises that is collected for the purpose of the study at the LISSI laboratory \cite{LISSIdataset}. The experimental results show that the final adapted classifier is always able to recognize with a good rate the human activities using their high-dimensional feature vectors. Besides, the results illustrate the superiority of the proposed model over other state-of-the-art classification methods in terms of of robustness w.r.t. weighted F1-score.
\end{itemize}

From this study, it has been concluded that the proposed approach can be applicable to other transfer learning problems as well.

The overall structure of this paper takes the form of six sections, including this introductory one.
Section \ref{related_work} gives a brief overview of state of the art concerning domain adaptation.  The proposed model is presented in section \ref{proposed_model}.
The fourth
section is laying out the experimental setup. The next section is dedicated to experiments, results and analysis of the evaluation. Finally, section \ref{future_works} includes a discussion of the findings and future research into this area.

\section{Related Work}
\label{related_work}
Reducing the gap between source and target domains in machine learning in order to make machine learning algorithms able to self adapt their models to the target domain, has been widely investigated these latest years. The approaches proposed in the state of the art are classified according to three main categories, namely  sample-based, feature-based, inference-based approaches \cite{Kouw2019}. The first one exploits the source distribution and target the minimization of the target risk estimation without target labels. The second category uses techniques for matching the data distributions, shifting of data, sub-space mapping, etc, to make transformations that maps source data onto target data. The third category targets the adaptation of the inference procedure by including constraints related to the target domain in the source model, incorporating uncertainties through Bayesian inference, etc. Semi-supervised, transfer learning, deep learning, adversarial learning  are representative examples of techniques used in the HAR approaches.

An unsupervised source selection algorithm was proposed in \cite{wang2018deeptransfer}. This algorithm is able to select the most similar K source domains from a list of available domains. Next, an effective Transfer Neural Network performs knowledge transfer for Activity Recognition (TNNAR) by capturing both the time and spatial relationship between activities during transfer. TNNAR was evaluated by body-part translation experiments, which provides more amount of samples for the model while the samples were limited to 4 common classes of activity on each dataset. Therefore, the effectiveness of TNNAR is not guaranteed on the imbalanced small-size datasets.

Authors in \cite{hossain2017active} investigated different active learning strategies to scale activity recognition and proposed a dynamic k-means clustering based active learning approach. Using active learning alleviates the labeling effort in the data acquisition and classification pipeline.
In spite of all improvement provided in the proposed model, such as computational complexity mitigation, the methods is yet prone to under-fitting down to its limited generalization ability \cite{wang2018deeptransfer}.

Pan et al. introduced Transfer Component Analysis (TCA), for domain adaptation which learns transfer components across domains in a reproducing kernel Hilbert space using maximum mean discrepancy \cite{pan2011domain}. The subspace spanned by these transfer components, preserves data properties and distributions of different domains close to each other. Therefore, using the new representations in this subspace, we can apply standard machine learning methods to train classifiers or regression models in the source domain and test it in the target domain. It means TCA applies representation transfer, and the classification task should be learned in another step. Besides, TCA only learns a global domain shift and does not fully consider the intra-class similarity \cite{wang2018stratified}.

The Geodesic Flow Kernel (GFK) presented in \cite{gong2012geodesic}, models domain shift by integrating an infinite number of subspaces that characterize changes in geometric and statistical properties from the source to the target domain. GFK model learns feature representations that are invariant across domains.

Some approaches perform transfer learning by reweighing or taking samples of the source domain. Balanced Distribution Adaptation (BDA) is proposed in \cite{wang2017balanced}, which adaptively leverages the importance of the marginal and conditional distribution discrepancies. Based on BDA, a novel Weighted Balanced Distribution Adaptation (W-BDA) algorithm was proposed to tackle the class imbalance issue in transfer learning by considering not only the distribution adaptation between domains but also adaptively changes classes' weights.

TransAct is another transfer learning-enabled activity recognition model introduced in \cite{khan2017transact} mitigates the degrade of recognition performance confront with activities with limited labeled samples. It addressed the challenges by augmenting the Instance-based Transfer Boost algorithm with k-means clustering. This model is designed only to compensate for the domain shift in activity level.

Authors of \cite{wang2018stratified}, proposed a general cross-domain learning framework that can exploit the intra-affinity of classes to perform intra-class knowledge transfer named Stratified Transfer Learning (STL). First, it obtains pseudo labels for the target domain by majority voting technique. Then, it performs intra-class knowledge transfer iteratively to transform both domains into a common subspace. The model was extended to accomplish both source selection and knowledge transfer later in \cite{chen2019cross}.
Although their research is dedicated to the HAR domain, the evaluation is limited to adaptation of body parts on the same person or similar body parts on different person. Besides, both TransAct and STL model utilized time domain and frequency domain feature extraction as their input which is not suitable for recognition of high-level activities with complicated patterns.

Since the introduction of Generative Adversarial Networks (GANs) \cite{goodfellow2014generative}, adversarial machine learning is gaining increasing attention and achieving today impressive performance in a wide variety of domains such as medicine \cite{nie2017medical,schlegl2017unsupervised}, text and image processing \cite{reed2016generative, zhang2017stackgan} and architecture \cite{bidgoli2019deepcloud, ueda2018data}. The idea behind GANs is putting generator and discriminator algorithms against each other in order to differentiate between the generated samples and real-world samples.
 Deep learning is used to build discriminators that continuously learn the best set of features making it hard for the generator to pass the discriminator test \cite{SenseGen2017}.

 The preliminary attempts of applying adversarial machine learning for HAR tackled the problem of producing synthetic data. In the latter, ground-truth annotations are generated automatically and the generator has complete knowledge about the target system. \cite{saeedi2018personalized, wang2018sensorygans}. Nonetheless, enhancing the classification methods remains the main important challenge in this field.

With respect to knowledge transfer, some attempts applied adversarial machine learning for training a robust deep network by way of a common representation for knowledge transfer \cite{tzeng2017adversarial, luo2017label}. Benefiting from the generalized adversarial framework, the generator component can be independently used as the target representation in order to reduce the difference between the training and test domain distributions and thus improves generalization performance.
Bousmalis et al. proposed a generative adversarial network (GAN)-based method that adapts source domain images to appear as if drawn from the target domain \cite{bousmalis2017unsupervised}. Having transferred samples by this instance-based adaptation method, machine learning models can be trained over the target domain. SA-GAN is another adversarial instance-based transfer learning approach introduced in \cite{Soleimani2019Cross} which provides enough data to train a classifier on the target domain.

Despite all the efforts dedicated to developing Adversarial Domain Adaptation approaches for HAR, the majority of the attempts are exploiting only data produced by vision sensors. 
The approaches exploiting wearable sensors such as accelerometers have a high tendency of being trapped by mode collapse problem, due to the continuous nature of the sensor's data.
Besides, they are not immune to the imbalanced class distribution problem; thus, the classes with lower probability density value may have less chance of being generated.

\section{Method}
\label{proposed_model}

\subsection{Objective}
Sensor-based HAR can be formulated as predicting current activity according to a sequence of sensors outputs $x_{i}$ \cite{wang2019deep}:
\begin{equation}
\label{har_def}
   f: W \rightarrow Y \mid W = \{x_{i} : i =1,...,n\}
\end{equation}
Whereas the correct activity sequence (ground truth) and input window of size $n$ denoted as $Y$ and $W$ respectively.
Any HAR related dataset has a finite amount of samples that are obtained from a limited number of human subjects. 
However, considering the requirements of applying HAR in real-world conditions, it is more interesting to evaluate the performance of a HAR model against many human subjects who their behaviors' data have not been included in the training dataset.

Let us define a learning domain $D$ and a learning task $T$ as bellow \cite{pan2010survey}:

\begin{equation}\label{domain_def}
  D = \Big(X, P(X)\Big) , T = \Big(Y, P(Y|X)\Big)
\end{equation}

Where feature space $X$ follows the distribution $P$ and $P(Y|X)$ formulates conditional distribution of label space $Y$. The shift between the domains may root in the learning domain, learning task, or both.
Furthermore, the source and the target domains may be dissimilar also in terms of class distributions, typically known as \textit{class imbalance} problem in machine learning. The conditional distributions of feature values are the same in source and target domains, yet the labels may not follow the same distribution in both domains.

\begin{figure*}[t]
    \setlength{\fboxrule}{0pt}
    \framebox{
      \parbox{0.95\textwidth}{
        \centering
        \includegraphics[width=0.95\textwidth, height=10cm, keepaspectratio]{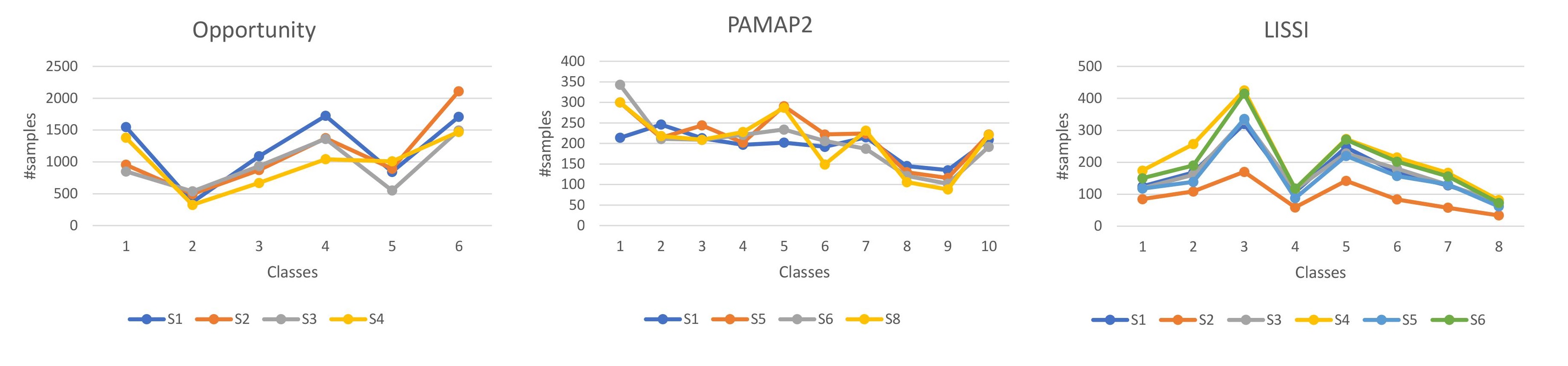}}}
        \caption{The class distribution of each subject P(Y) in Opportunity, PAMAP2 and LISSI dataset.}
        \label{dataset_infogram}
\end{figure*}

Subject level domain adaptation concentrates on the generalization of the knowledge a machine learning model. The latter is trained from a known subject and should be extended to unknown or unseen subject. Let us consider an HAR system that is supposed to recognise the activities of the inhabitants of smart homes. The inhabitants are considered as new subjects from the perspective of the HAR system. Even if the HAR system can  be setup to temporarily collect data and learn inhabitants activities in the same time in a kind of a system initialisation mode, the annotation of the collected samples, by human experts or the inhabitant themselves  to apply supervised machine learning will be infeasible. In this case the appropriate approach is a semi-supervised learning, where labeled data is provided in the source domain (subject) while target domains' samples are label excluded. Formally, the objective is to adapt the target domain $D_{t} = \big(X_{t}, P(X_{t})\big)$ to the source domain $D_{s} = \big(X_{s}, P(X_{s})\big)$ so as to have enough labeled data to train a HAR model on target domain. It has commonly been assumed that in the class imbalance domain adaptation problems, $P(X_s|Y_s = y_i) = P(X_t|Y_t = y_i)$ is held for all classes $i$, though the distribution of classes may not be the same in both domains \cite{jiang2008literature}. Fig. \ref{dataset_infogram} exemplifies this concept in the datasets used for this study. Taking any pair of subjects as the source and target domain, there exists a shift between the class probability distribution of both domains, which means $P(Y_s) \ne P(Y_t)$.

Classification of data with imbalanced class distribution along with absence of the labels in the target domain, may pose a significant drawback of the performance attainable by the adaptation process.

\subsection{Adversarial Adaptation}
\label{adv_adap}
Domain adaptation methods can be sorted out into 4 categories based on the type of knowledge transferred: Instance, feature representation, parameter, and relational transfer \cite{Soleimani2019Cross}. The first two categories focus on drawing the samples of both domains closer, by direct transformation or finding a common representation, respectively. Parameter and relational transfer methods transform prior knowledge and parameters and data relationship between domains. The proposed solution is a kind of instance transfer except that it combines the data transformation and classifier training procedure.

More formally, let us consider $X^{s} = \{(x_{s}, y_{s})^{i} \mid i=0 \to n_{s}\}$ represents the set of $n_{s}$ labeled samples from the source domain $D_{s} = (X_{s}, P(X_{s}))$ and $X^{t} = \{(x_{t})^{i} \mid i=0 \to n_{t} \}$ denotes the set of $n_{t}$ unlabeled samples from the target domain $D_{t} = (X_{t}, P(X_{t}))$.
The proposed adversarial adaptation model consists of a Generator (G), Discriminator (D), and Classifier (C). The generator $G(x, z; \theta_G)$ is a differentiable function represented by a Convolutional Neural Network (CNN) that generates synthetic data, called also fake samples, by using the input and noise vector. The discriminator $D(x; \theta_D)$ is defined as a CNN that outputs a single scalar indicating the probability that $x$ came from the target domain rather than the  generator. The classifier $C(x; \theta_C)$ is also a CNN predicts the class of the input.
These elements are playing a min-max game together based on the cost function $\mathbb{J}$ which combines the loss functions of adversary and classification tasks as follows \cite{Soleimani2019Cross}:
\begin{equation}
\label{eq:eq_overal}
    \begin{split}
       & \mathbb{J}(G, C, D, X^{s}, X^{t}) = \\ & \min_{G,C} \max_{D} \mu \mathbb{J}_{D}(D, G, X^{t}, X^{s})   + \lambda \mathbb{J}_{C}(C, G, X^{s})
    \end{split}
\end{equation}
The impact of the classification $\mathbb{J}_{C}$ and adversary $\mathbb{J}_{D}$ task loss on the generation task are reflected and controlled by $\mu$ and $\lambda$ coefficient respectively in $\mathbb{J}$.
Generator supposed to generate artificial data which are similar to the samples of the target domain by descending on the gradient of $\mathbb{J}(G, C, D, X^{s}, X^{t})$:
\begin{equation}
\label{G_training}
\begin{split}
& \frac{\partial}{\partial \theta_{G}}  \mathbb{J}(G, C, D, X^{s}, X^{t})= \\ & \mu \nabla_{\theta_{G}} \mathbb{J}_{D}(D, G, X^{t}, X^{s})   + \lambda \nabla_{\theta_{G}} \mathbb{J}_{C}(C, G, X^{s})
\end{split}
\end{equation}
During the training phase, these fake generated samples are getting as similar as possible so that the discriminator will not be able to discriminate them from the original target data.

The training samples are commonly feeding into the adversarial frameworks in mini-batch form, to avoid the mode collapse problem in which, the generator learns only to generate fake samples from a few classes (modes) of the data distribution, albeit the samples from the ignored modes appears in the training set \cite{srivastava2017veegan}.
Therefore, the generator collapses into the few modes that discriminator assumes them highly realistic.
On the contrary, feeding the samples in mini-batch to the discriminator rather than in isolation, gives a broader horizon to the discriminator and possibly avoids the mode collapse.
However, it would not be practical enough in all cases, especially for the highly imbalanced small-size datasets such as LISSI. Mini-batch selection of the dataset would even aggravate the problem in practice since it scales down the sample size. It is highly probable that the less populated classes be left without a representative in some batches, as the sample size shrinks and the proportions of the classes in the sample space cannot be taken for granted. Consequently, the discrimination of that certain classes would be unconcerned in the gradient computation of the batch due to the disappearance of their samples.

\subsection{Proposed method}

To tackle this issue, we applied a \textit{micro-mini batch} strategy of learning. Each mini-batch consists of $\mathbb{C}$ micro-batches whereas $\mathbb{C}$ is the number of classes of activities in the dataset. Micro-batches contain $m$ samples randomly drawn without replacement from each one of $\mathbb{C}$ classes while $m$ can be set by the value of the least populated class population in the dataset.
In this way, the proposed adversarial approach does not only prevent the Mode Collapse and its related issues but also enables a kind of parallelism of instance transfer and re-training of the target domain's classifier.

The optimization problem for discriminator can be solved by ascending the gradient of a Mean Squared Error (MSE) loss function $\mathbb{J}_{D}$:
\begin{equation}\label{D_training}
\begin{split}
& \mathbb{J}_{D}(D, G, X^{t}, X^{s}) =
      \frac{1}{ m\mathbb{C}}\mathlarger{\sum_{j=1}^{\mathbb{C}}
       \sum_{i=1}^{m} \Big[}  \Big(1 - D\big( x_{t}^{(i)}\big) \Big)_{x_t^{(i)}\in b_t^{j}}^2 \\ & +  \Big( 1+ D\big(G(x_{s}^{(i)}, z)\big) \Big)_{x_s^{(i)}\in b_s^{j}}^2 \mathlarger{\Big]}
\end{split}
\end{equation}
where $b_{s}^{j}$ and $b_{t}^{j}$ refer to the j-th micro part of the current mini-batch from the source $X^{s}$ and target $X^{t}$ domain samples, respectively. The ground truth for real/fake indication is considered as $\pm1$ and $z$ denotes a random noise vector.

Correspondingly, the classifier attempts to assign a right label to its inputs including source domain data and the synthetic data generated by optimizing the cross-entropy loss function $\mathbb{J}_{C}$:
\begin{equation}\label{C_training}
\begin{split}
 &\mathbb{J}_{C}(C, G, X^{s}) =  \frac{1}{m\mathbb{C}} \mathlarger{\sum_{j=1}^{\mathbb{C}} \sum_{i=1}^{m}}\\&
  \Big[  -y_{s}^{(i)} \log C\big( x_{s}^{(i)}\big) - y_{s}^{(i)}\log \Big( C\big ( G(x_{s}^{(i)}, z ) \big) \Big) \Big]_{x_s^{(i)}, y_s^{(i)}\in b_s^{j}}
\end{split}
\end{equation}
Both the discriminator and classifier components get updated based on their gradient $\frac{\partial}{\partial \theta_{D}} \mathbb{J}_D$ and $\frac{\partial}{\partial \theta_{C}} \mathbb{J}_C$, yet any gradient-based optimizer can be exploited for each component independently.
Finally, when the training loss values converge, the training phase can be terminated and the classifier component will be functional independently.

Algorithm \ref{Algorithm} outlines the training steps. In summary, each iteration of the training procedure consists of 3 steps for the mini-batch update of discriminator D, classifier C, and generator G, respectively. Reordering the steps may affect convergence flow. The components of the model together struggle to close in the distribution of target domain samples on those of source domain where the labels are available. Having samples with approximately the same distribution, source domain labels are compatible to be exploited in supervised training of the classifier. As the distribution of generated samples $P(G(X^{s}, z))$ getting closer to $P(X^{t})$, the performance of the classifier C improves since the source labels more deeply cohere with target inputs.

\begin{algorithm*}[ht]
\caption{
    {Micro mini-batch training of the proposed model}
}
\KwIn{$X^{s}, X^{t}, Y^{s}$}
\KwOut{C}
\BlankLine

\For{number of training iterations or until convergence}{
    \For{number of mini-batches of data}{
    \begin{enumerate}
      \item Sample a micro-batch of size $m$ from $X^{s}, X^{t}, Y^{s}$ of class $j = 1, ... , \mathbb{C}$ of data:\\
      $b_s^j=\big\{(x_{s}^{(1)}, y_{s}^{(1)}), (x_{s}^{(2)}, y_{s}^{(2)}), ... , (x_{s}^{(m)}, y_{s}^{(m)})\big\}$,
      $b_t^j=\big\{x_{t}^{(1)}, x_{t}^{(2)}, ... , x_{t}^{(m)}\big\}$

      \item Make a mini-batch of source $\{b_s^1, b_s^2, ... , b_s^\mathbb{C}\}$ and target domain $\{b_t^1, b_t^2, ... , b_t^\mathbb{C}\}$ samples \\

      \item  Update discriminator D by ascending its stochastic gradient:\\
          \begin{center}
             $ \nabla_{\theta_{D}} \mathlarger{
              \frac{1}{ m\mathbb{C}}}\mathlarger{\sum_{j=1}^{\mathbb{C}}
       \sum_{i=1}^{m} \Big[}  \Big(1 - D\big( x_{t}^{(i)}\big) \Big)_{x_t^{(i)}\in b_t^{j}}^2 +  \Big( 1+ D\big(G(x_{s}^{(i)}, z)\big) \Big)_{x_s^{(i)}\in b_s^{j}}^2 \mathlarger{\Big]} $
          \end{center}

      \item Update classifier C by ascending its stochastic gradient:\\
          \begin{center}
             $\nabla_{\theta_{C}} \mathlarger{\frac{1}{m\mathbb{C}}} \mathlarger{\sum_{j=1}^{\mathbb{C}} \sum_{i=1}^{m}}
  \Big[  -y_{s}^{(i)} \log C\big( x_{s}^{(i)}\big) - y_{s}^{(i)}\log \Big( C\big ( G(x_{s}^{(i)}, z ) \big) \Big) \Big]_{x_s^{(i)}, y_s^{(i)}\in b_s^{j}}
  $
          \end{center}

      \item Update generator G by ascending its stochastic gradient:\\
             $\nabla_{\theta_{G}} \mathlarger{ \frac{1}{m\mathbb{C}} \sum_{j=1}^{\mathbb{C}} \sum_{i=1}^{m}}\Bigg[
             \lambda \Bigg(-y_{s}^{(i)} \log C\big( x_{s}^{(i)}\big)- y_{s}^{(i)}\log \Big( C\big ( G(x_{s}^{(i)}, z) \big) \Big) \Bigg) +  \mu \Bigg( \Big(1 - D\big( x_{t}^{(i)}\big)\Big)^2 +  \Big( 1+ D\big(G(x_{s}^{(i)}, z)\big) \Big)^2 \Bigg)
             \Bigg]$
             \\
             \centering ${x_s^{(i)}, y_s^{(i)}\in b_s^{j}, x_t^{(i)}\in b_t^{j}}$

      \end{enumerate}
    }
}
Return classifier C.

\label{Algorithm}
\end{algorithm*}

\subsection{Adversarial Training Techniques}
Training GANs includes obtaining a Nash equilibrium to a two-player non-cooperative game while each player wishes to minimize its own cost function \cite{salimans2016improved}. A Nash equilibrium for our problem is a triple $(\theta_{D}, \theta_{G}, \theta_{C})$ such that $\mathbb{J}_{D}(G, D, X^{s}, X^{t})$ is at the maximum with respect to $\theta_{D}$, and  $\mathbb{J}(G, C, D, X^{s}, X^{t})$ and $\mathbb{J}_{C}(C, G, X^{s})$ are at the minimum with respect to $\theta_{G}$ and $\theta_{C}$, respectively. Thought, finding Nash equilibria is quite problematic since maximizing $\mathbb{J}_{D}$ contradicts the minimization of two remaining cost functions. In addition, a very confident discriminator pose several challenges to adversarial training such as vanishing gradients phenomena on the generator, as well as  instability of the generator gradient updates \cite{arjovsky_towards_2017}. Different techniques have been utilized in order to overcome these challenges \cite{salimans2016improved,arjovsky_towards_2017}. The mini-batch discrimination strategy applied to avoid the mode collapse failure by allowing the discriminator to look at multiple samples in combination, rather than in isolation.

Label smoothing is another technique that replaces the 0 and 1 targets for a classifier with smoothed values, such as 0.9 or 0.1. Replacing positive class labels with $\alpha$ and negative class labels with $\beta$, the optimal discriminator is formulated as the following:
\begin{equation}
\label{eq:eq_optimal}
    D^{*}(x) = \frac{\alpha P(X_{t}=x) + \beta P(X_{g}=x)}{P(X_{t}=x) + P(X_{g}=x)}
\end{equation}
Adding continuous noise to the inputs of the discriminator can fix the instability and vanishing gradients issues and smoothen the distribution of the probability mass. Therefor, the optimal discriminator is re-written as follows:
\begin{equation}
\label{eq:eq_optimal_noisy}
    D^{*}(x) = \frac{\alpha P(X_{t}= x + \epsilon) + \beta P(X_{g}=x  + \epsilon)}{P(X_{t}= x + \epsilon) + P(X_{g}=x + \epsilon)}
\end{equation}
Whereas $\epsilon$ denotes a continuous random variable with density $P_{\epsilon}$. In our case,  we chose the noise inputs from uniform distribution within the interval [0, 1).

\section{Experimental Setup}
\label{experimental_setup}
\subsection{Dataset}
To evaluate the performance and robustness of the proposed model, we selected three representative datasets of large, medium, and small sizes. Several HAR-related datasets are publicly available, though they mostly have their samples distributed into many subjects. The goal is to take each subject as a domain and transfer the knowledge among them. Hence, we consider the average sample size per each subject in the datasets to measure the size and select appropriate datasets based on it.
The experiments have been conducted on Opportunity and PAMAP2 benchmark datasets \cite{chavarriaga2013opportunity,reiss2012introducing} and the LISSI dataset  \cite{LISSIdataset}. The latter is collected for the purpose of the study at the LISSI laboratory. Those datasets are briefly introduced in the following lines.
\begin{itemize}
    \item PAMAP2:
Totally 9 subjects were participating in the data acquisition of the PAMAP2 dataset following a protocol of 12 activities and 6 optional ones while wearing 3 IMUs and a Heart Rate-monitor \cite{reiss2012introducing}. Among them, 4 subjects with the more well-distributed classes of samples have been selected for evaluation.
    \item Opportunity:
It consists of wearable sensors's output while there were worn by 4 human subjects who were performing predefined unscripted daily living activities \cite{chavarriaga2013opportunity}. Four groups of activity are defined in this dataset based on the abstraction level. Generally, the more abstract activities may have more complicated and diverse patterns. To exemplify, let us compare the high level \textit{Coffee Time} to the low level \textit{Sit} locomotion activity. Since the activities are not scripted, there are different attitudes to make and drink coffee. Therefor, the related patterns would be more challenging to be recognized by machine learning techniques. Opportunity dataset contains 5 classes of high level activities, we aim to recognize in this research.
    \item LISSI:
This dataset has been collected in the context of the Medolution project in order to develop HAR methods that can be applied for the remote monitoring of the rehabilitation self exercises performed at home. The dataset consists of sensors' data of human subjects, which repeated a complete sequence of five rehabilitation self exercises. The subjects are equipped with wearable inertial sensors and follow  a video tutorial in order to perform the exercises in the sequencing indicated by the tutorial. The data acquisition platform comprises 5 wearable sensors (Xsens Inertial Measurement Units) and 3 Kinect RGB-D cameras, which can capture the body movements from three perspectives: front, profile and the head. The LISSI dataset consists of 8 high-level classes corresponding to the self exercises, each one encapsulates a series of low-level classes, corresponding to the self exercise components, also called low-level exercises. The latter are composed of repetitive or non-repetitive movements, which may occur multiple times during the high-level exercise. The dataset includes also contextual labels that indicates how the self exercise have been performed. The dataset has been collected during two different periods involving respectively 7 and 11 subjects. Only the data belonging to 6 human subjects involved in first period are have been annotated and used for the purpose of the present study.
\end{itemize}

\begin{table*}[ht]
\caption{Hyper-parameters of the proposed adversarial model's components}
\label{elements_structure}
\begin{center}
\resizebox{\textwidth}{!}{
\begin{threeparttable}
\begin{tabular}{l|l|l}
\hline
\textbf{Element} & \textbf{Parameter}              & \textbf{Values}          \\
\hline \hline \\[-1.5ex]
           & Input layer dimension           & (100,), (100,), (4000,)\tnote{\dag}  \\
           &                                 &                        \\
           & Number of convolutional blocks       & $cb$                   \\
           & Kernel size of $1^{st}$, $2^{end}$ Conv.  layers in convolutional blocks  & 3, 3  \\
Generator  & Sliding stride of $1^{st}$, $2^{end}$ Conv. layers in the blocks & 1, 1  \\
           & Number of filters in $1^{st}$, $2^{end}$ Conv. layers in the blocks & $gf, gf$ \tnote{*}\\
           &                                                                      &     \\
           & Number of filters of output Conv. layer                        & (100,) , (100,), (4000,)\tnote{\dag}   \\
           & Kernel size of output Conv. layer                                    & 3   \\
\hline  \\[-1.5ex]
           & Input Layer dimension                                          &(100,) , (100,), (4000,) \tnote{\dag} \\
           &                                                                &        \\
Classifier & Number of filters for 1, 2, $3^{th}$ Conv. layers           & $cf, cf/2, cf/4$ \tnote{*}\\
           &                                                                &        \\
           & Number of neurons in Dense output layer                                & 6, 8, 10 \tnote{\dag} \\
\hline  \\[-1.5ex]
           & Input Layer dimension                                          & (100,) , (100,), (4000,)\tnote{\dag} \\
           &                                                                &        \\
Discriminator & Number of filters for Conv. layers           & $2\times df, 4\times df, 8\times df, 4\times df, 2\times df$ \tnote{*}\\
           &                                                                &        \\
           & Number of the units in the output Dense layer                  & 1      \\

\end{tabular}
\begin{tablenotes}
\item[\dag] \textit{for Opportunity, LISSI, and PAMAP2 datasets, respectively.}
\item[*] \textit{the variablea $gf$ (generator filters), $cf$ (classifier filters), and $df$ (discriminator filters) should be tuned in training.}
\end{tablenotes}
\end{threeparttable}
}

\end{center}
\end{table*}

\begin{figure}[ht]
    \setlength{\fboxrule}{0pt}
    \framebox{
      \parbox{0.95\columnwidth}{
        \centering
        \includegraphics[width=0.95\columnwidth, height=10cm, keepaspectratio]{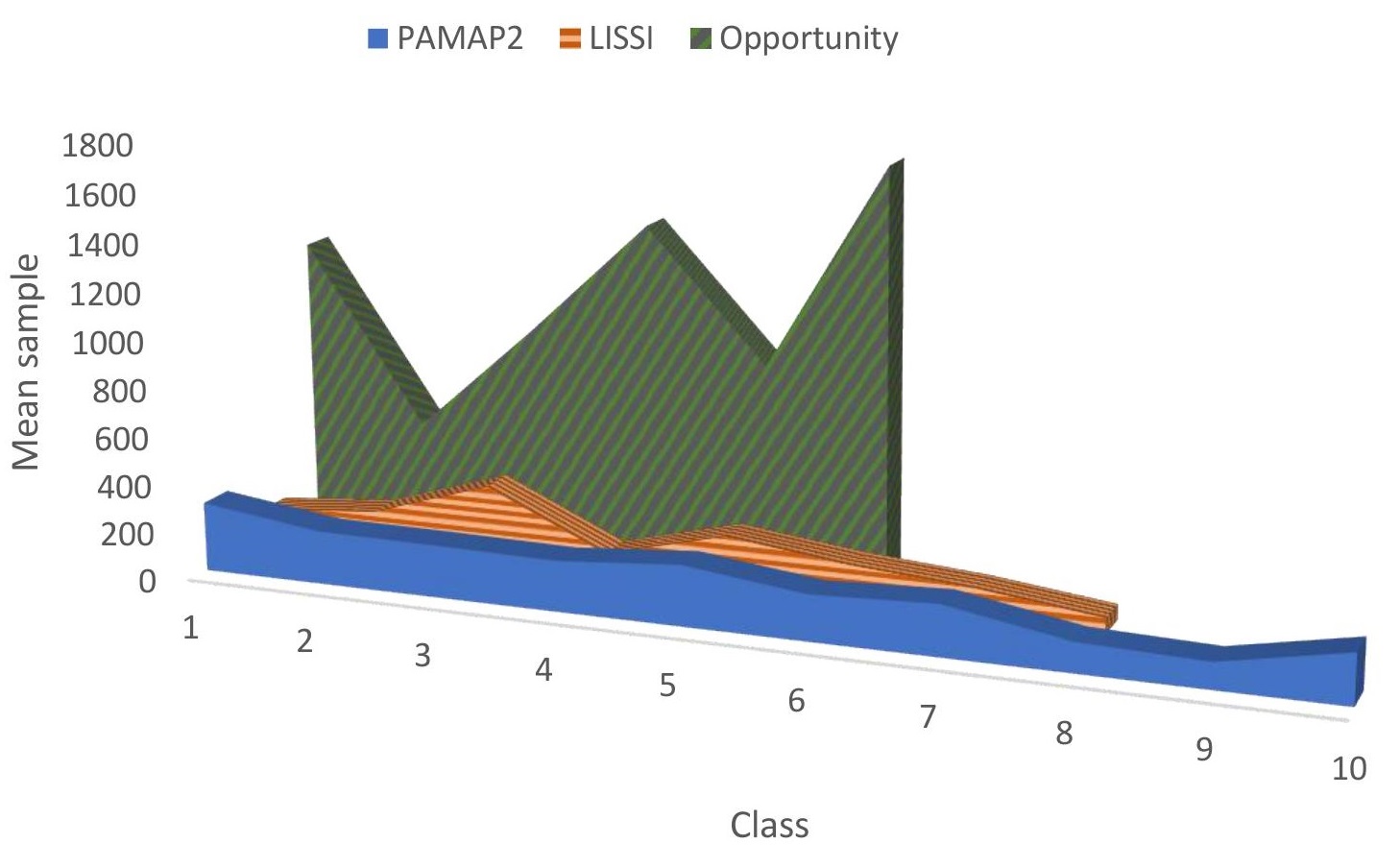}}}
        \caption{Distribution of the average number of samples of each class for subjects in Opportunity, PAMAP2, and LISSI datasets.}
        \label{imbalanced_dist}
\end{figure}

\subsection{Data Preparation}
Data preparation for all the 3 datasets includes replacing missing values, data segmentation and normalization. A min-max normalization has been applied based on the sensor’s range of each dataset. Dimension reduction was applied for Opportunity and LISSI datasets.

Approximately 3, 1.5 and 5 seconds length sliding window was employed to segment data for Opportunity, LISSI and PAMAP2 dataset respectively, taking into account 70\% of overlap between consecutive windows of data.
Although segmenting the samples into windows, provides more information about ongoing activity for the system, it enlarges the dimensionality of data to be processed.

To exemplify, let us consider 3 and 1.5 seconds windows of data which are collected with the frequency of 60 and 30 Hz for Opportunity and LISSI datasets, respectively.
Having feature size of 113 and 95, segmentation lead into input vectors with thousands dimensions. Handling high-dimensional data samples is way more laborious and time consuming for artificial neural network. Hence, dimension reduction technique may come as a boon to facilitate the training process. Using Principal Component Analysis (PCA), we scaled dimensions down to 1\% and 2\% for Opportunity and LISSI by investigating the precision-time trade-off. However, dimension reduction imposed a crucial decrease in the performance of classification task of PAMAP2 dataset. Therefore, this step is skipped for this dataset.

Statistics of LISSI dataset show it has lower \textit{Samples per Subject} rate while it has more subjects in comparison to the Opportunity dataset. Furthermore, There exist very short-length classes of activity such as \textit{Kneeling} or \textit{Warm up} which have fewer samples compared to the rest and make the dataset imbalanced. Since the research is focused on subject level adaptation, and a deep architecture is used in the proposed approach, we demand high amount of data for training.
We figured out this problem through the use of \textit{micro-mini-batch} learning strategy which is discussed in section \ref{proposed_model}.

\begin{figure*}[t]
    \setlength{\fboxrule}{0pt}
    \framebox{
      \parbox{0.98\textwidth}{
        \centering
        \includegraphics[width=0.98\textwidth, height=12cm, keepaspectratio ]{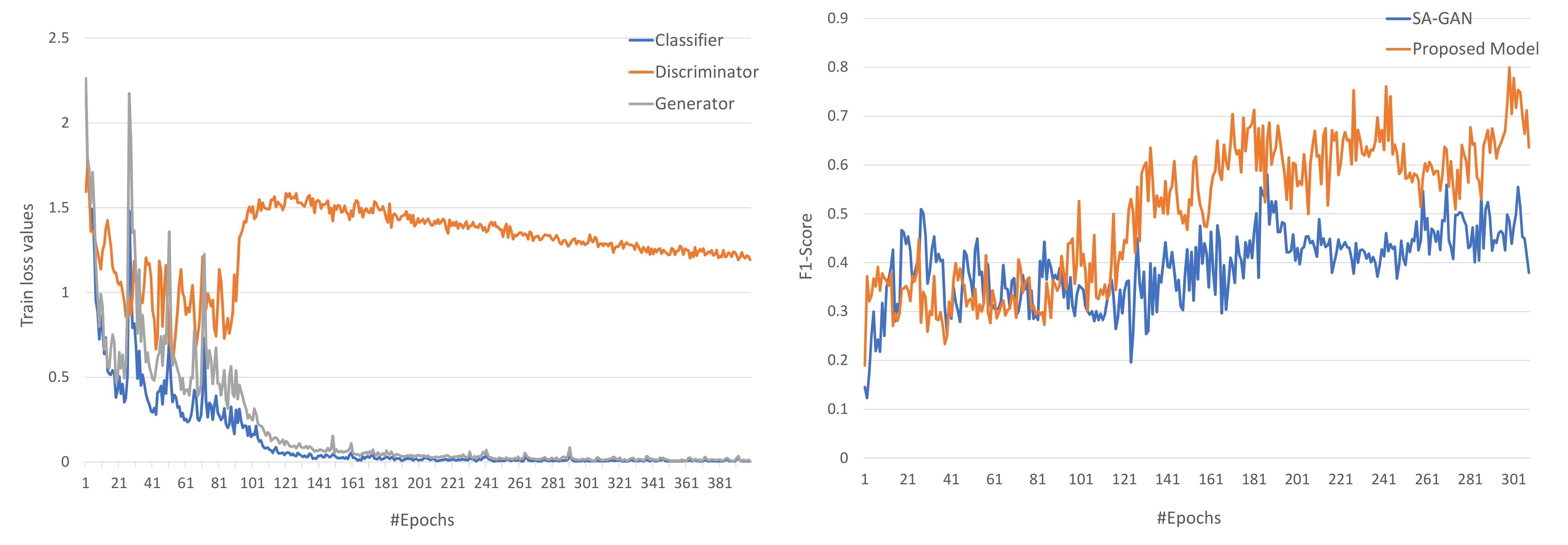}}}
        \caption{Left:Average loss values for the discriminator, generator and the classifier elements of the proposed model adversarial framework for translation of subject 5 to subject 6 in PAMAP2 dataset. Right: Accuracy of the classifier over the validation set of the target domain during the training of the framework to translate subject 5 to subject 1 of PAMAP2 dataset.}
        \label{train_loss_plot}
\end{figure*}

The datasets are split into training, validation and testing sets as follows. As for the Opportunity dataset, the samples of each subject were held in 5 Activity of Daily Living (ADL) ﬁles. The first 3 ADL files were dedicated to the training set, and the fourth and fifth ones considered for validation and testing sets, respectively. PAMAP2 and LISSI dataset, the samples of each subject divided into the training, validation and testing set proportional to 0.6, 0.1, 0.3, respectively. In summary, each experiment has a source and target subject which their samples are categorized into 3 disjoint subsets. The reported results obtained over the testing set for the target subject.

\section{Experiments}
\label{experiments}
As earlier discussed in section \ref{experimental_setup}, Opportunity, PAMAP2 and LISSI datasets are preprocessed for the evaluation of the proposed method. Set of subject translation experiments have been held to practically demonstrate the necessity of domain adaptation and its effectiveness as well.

For each dataset samples of each subject are considered independently as the source or target domain. For each target domain (subject) a classifier is adapted, examined and compared with two other adaptation methods, namely, Stratified Transfer Learning (STL) and Geodesic Flow Kernel (GFK) as well as adaptation performance upper bound. The implementations of all the experiments for STL and the proposed model have been done in Python using Keras and Tensroflow libraries. The codes of GFK and STL are provided in Matlab and can be obtained online\footnote{https://github.com/jindongwang/transferlearning/tree/master/code}.

Table \ref{elements_structure} details the components of the proposed approach. The complexity of the addressed problem is different for each source-target pair of subjects. Therefore, the complexity of the proposed method should be adjusted by using its controlling parameters, including $cb$, $gf$, $cf$, and $df$.
The high learning capacity of one-dimensional Convolutional Neural Networks, makes them more suitable for this application.
Additionally, employing CNN models facilitates the recognition task by extracting more abstract feature values rather than time and frequency based handcrafted feature extraction approaches.

\begin{table*}[ht]
\begin{center}
\caption{Comparison of the the generic proposed approach performance and GFK \cite{gong2012geodesic}, STL\cite{wang2018stratified} and SA-GAN\cite{Soleimani2019Cross}  model, in terms of Weighted F1 measure on \textbf{Opportunity} dataset. The most dominant performance in each transformation experiment marked in bold.}
\label{table_results_opportunity}
\centering
\resizebox{0.95\textwidth}{!}{
\begin{tabular}{ccccccccc}
Source Subject & Target Subject & Distance & No Transfer &  STL  & GFK  & SA-GAN & Proposed Model & Supervised \\
\hline \hline \\[-1.5ex]
\multirow{3}{*}{1} & 2     & 46.69     & 0.45     & 0.65      & 0.59   & 0.73  & \textbf{0.74}  & 0.75\\
                   & 3     & 45.10     & 0.27     & 0.37      & 0.43   & 0.45  & \textbf{0.58}  & 0.71 \\
                   & 4     & 77.15     & 0.40     & 0.47      & 0.55   & 0.49  & \textbf{0.57}  & 0.59\\
\hline \\[-1.5ex]
\multirow{3}{*}{2} & 1     & 40.47     & 0.48     & 0.52      &\textbf{0.62}   & 0.56  & 0.56   & 0.65 \\
                   & 3     & 34.38     & 0.44     & 0.46      & 0.51   &\textbf{0.52}  & 0.40   & 0.71  \\
                   & 4     & 72.80     & 0.29     &\textbf{0.46} & 0.40        & 0.39  & 0.42   & 0.59\\
\hline \\[-1.5ex]
\multirow{3}{*}{3} & 1     & 38.38     & 0.23     & 0.40      & 0.45   & 0.42    & \textbf{0.52} & 0.65\\
                   & 2     & 37.54     & 0.21     & 0.54      & 0.53   & \textbf{0.61}   & 0.52  & 0.75  \\
                   & 4     & 73.69     & 0.31     & 0.37      & 0.44   & 0.44    & \textbf{0.50} & 0.59   \\
\hline \\[-1.5ex]
\multirow{3}{*}{4} & 1     & 73.53     & 0.26     & 0.38      & 0.51   & 0.51   & \textbf{0.52}  & 0.65 \\
                   & 2     & 70.80     & 0.29     & 0.54      & 0.45   & 0.55   & \textbf{0.68}  & 0.75   \\
                   & 3     & 69.44     & 0.24     & 0.48      & 0.37   & 0.49   & \textbf{0.53}  & 0.71     \\
\end{tabular}
}
\end{center}
\end{table*}

\begin{table*}[ht]
\begin{center}
\caption{Comparison of the the generic proposed approach performance and GFK \cite{gong2012geodesic}, STL \cite{wang2018stratified} model, in terms of Weighted F1 measure on \textbf{PAMAP2} dataset. The most dominant performance in each transformation experiment marked in bold.}
\label{table_results_pamap2}
\centering
\resizebox{0.95\textwidth}{!}{
\begin{tabular}{ccccccccc}
Source Subject & Target Subject & Distance & No Transfer  & STL  & GFK & SA-GAN  & Proposed Model & Supervised \\
\hline \hline \\[-1.5ex]
       & 5     &  91.82   & 0.37 & 0.62 & 0.72 & 0.69 & \textbf{0.77}   & 0.98       \\
1      & 6     &  91.58   & 0.32 & 0.56 & 0.64 & 0.66 & \textbf{0.70}   & 0.97       \\
       & 8     &  107.07  & 0.04 & 0.57 & 0.49 & 0.65 & \textbf{0.72}   & 0.92       \\ \hline \\[-1.5ex]
       & 1     &  91.82   & 0.32 & 0.76 & 0.66 & 0.71 & \textbf{0.76}   & 0.99       \\
5      & 6     &  42.13   & 0.64 & 0.83 & 0.75 & 0.83 & \textbf{0.83}   & 0.97       \\
       & 8     &  56.01   & 0.26 & 0.52 & 0.69 & 0.66 & \textbf{0.73 }  & 0.92       \\ \hline \\[-1.5ex]
       & 1     &   91.58  & 0.16 & 0.67 & 0.56 & 0.61 & \textbf{0.78}   & 0.99       \\
6      & 5     &   42.13  & 0.41 & 0.74 & 0.75 & 0.79 & \textbf{0.83}   & 0.98       \\
       & 8     &   56.76  & 0.17 & \textbf{0.86} &  0.58 & 0.63 & 0.82   & 0.92       \\ \hline \\[-1.5ex]
       & 1     &  107.07  & 0.10  & 0.54 & 0.58 & 0.68 & \textbf{0.76}  & 0.99       \\
8      & 5     &  56.01   & 0.25 & 0.55 & 0.41  & 0.60 & \textbf{0.73}  & 0.98       \\
       & 6     &  56.76   & 0.27 & 0.58 & 0.61  & \textbf{0.73} & 0.71  & 0.97       \\
\end{tabular}
}
\end{center}
\end{table*}

\subsection{Results and Analysis}
We evaluate the proposed model by conducting extensive experiments on three datasets of different size to assess its functionality and robustness. We opted for W-F1 measure as the evaluation metric since it gives better insight comparing to accuracy, precision, and recall deliberating imbalance distribution of classes in the dataset. Fig. \ref{imbalanced_dist} depicts the mean sample rate for each class on these datasets. Opportunity dataset is the most populated one among the selected datasets; nonetheless, it supposed to be challenging due to its highly imbalanced distribution. PAMAP2 dataset includes the smoothest distribution of classes; however, it contains more classes of activities with less sample/class rate. Finally, LISSI dataset is expected to be the most challenging one owing to its less populated and imbalanced distribution of samples.

\textbf{Analysis of the model during the training}: Fig. \ref{train_loss_plot} portrays the loss value trends for three main components of the proposed model during a translation experiment on the PAMAP2 dataset. The plot highlights the adversary of generator and discriminator as it was presumed. The loss values for $D$ and $G$ are fluctuating oppositely; in the periods that $G$ is improving by decreasing its loss values, $D$ is deteriorating and vice versa. The whole framework converges after about 150 epochs in this experiment, when the discriminator achieves desired uncertainty($D^*$), generator and classifier both converges and no further improvement can be reached by optimizing the loss functions.
The trend of accuracy improvement of the proposed model and SA-GAN in Fig. \ref{train_loss_plot}, approves the effectiveness of our proposed model which improves the performance of adversarial adaptation. The plot compares the accuracy of the models over the validation set of the samples from the target subject in PAMAP2 dataset. Although, both models set out similar performance in the beginning, the proposed model shows more superiority over SA-GAN after 100 epochs as it is reaching closer to the convergence.

\begin{figure*}[t]
    \setlength{\fboxrule}{0pt}
    \framebox{
      \parbox{0.95\textwidth}{
        \centering
        \includegraphics[width=0.95\textwidth, height=9cm, keepaspectratio]{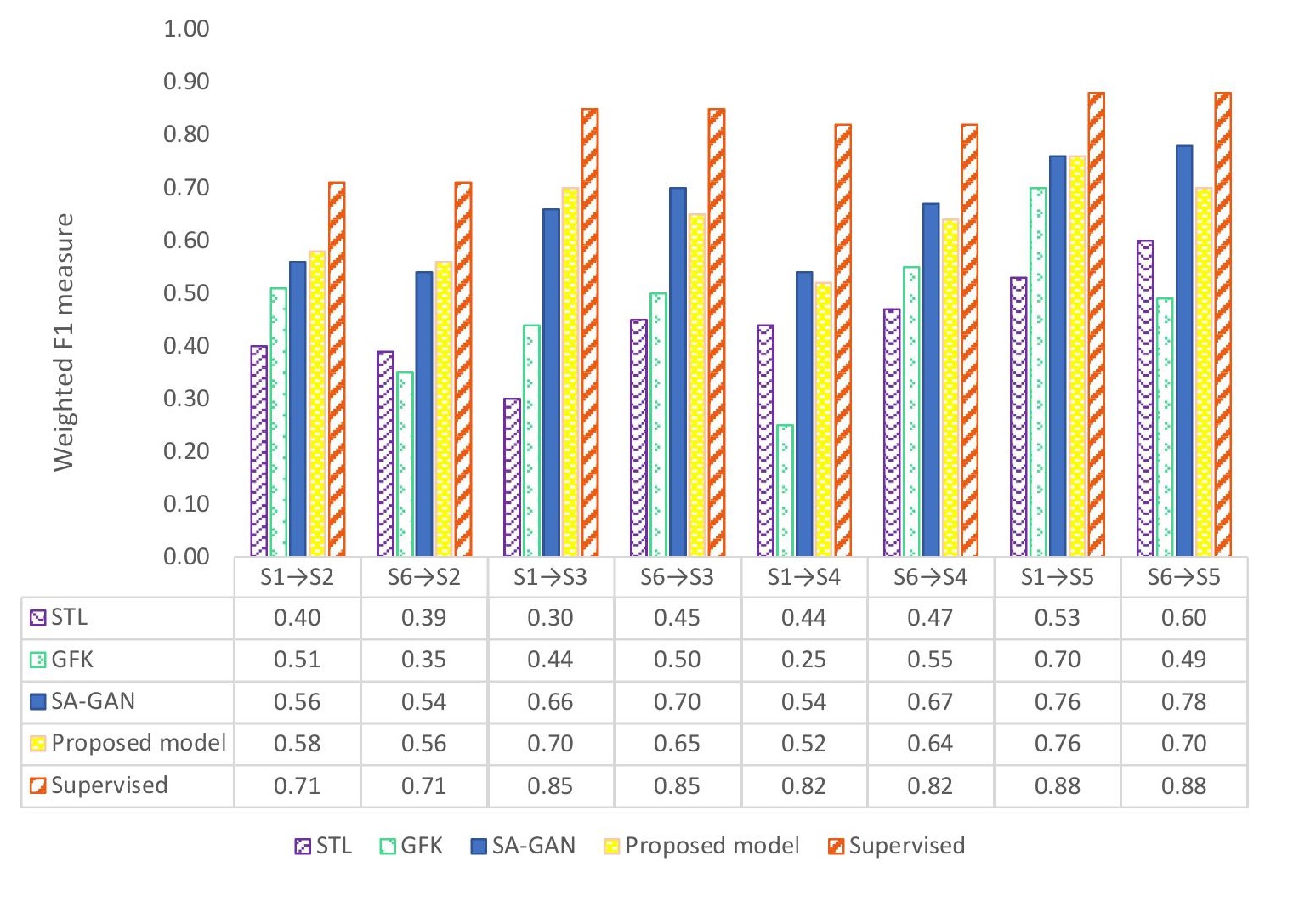}}}
        \caption{Results of applying Subject to Subject Transfer Learning on \textbf{LISSI} dataset.}
        \label{lissi_resultstraining_plot}
\end{figure*}

\textbf{Evaluation of the model over different datasets and comparison with state-of-the-art (test phase):}
Table \ref{table_results_opportunity} outlines the results of subject-level transfer learning on Opportunity dataset comparing with other states of the art models. This dataset is considered a large scale one since it holds on average 6400 windows of samples per subject. The \textit{No Transfer} column refers to the experiments have been done to justify the necessity of domain adaptation by training a Convolutional Neural Network over the source domain and testing it against the target domain samples, directly without any adaptation technique.
Since the proposed method is a semi-supervised one, we considered the outcomes of the CNN trained under the supervision, as the upper bound for domain adaptation. Training a model with the annotated data from the same domain, is considered as the highest margin that can be accomplished. Hence, the adaptation performance regarded as convincing as it is close to the upper bound.
We find that our proposed method improved the performance of classification task in compare with other adversarial domain adaptation method SA-GAN by 7.3\% on average which admits the effectiveness of micro-mini-batch training approach. Entirely, the proposed model improves the classification performance in 8 out of 12 experiments, accomplished best and second-best W-F1 measure in more than 83\% of the experiments.

The results on Opportunity dataset, concur the competence of the proposed approach in large scale datasets. Let us consider all three available source of knowledge transfer for each source domain in this dataset. The classifier adapted by the proposed model presented the best classification results for 3 out of 4 target domains and  competitive result for the remaining one.

Table \ref{table_results_pamap2} summarizes the results of classification on a medium-size dataset, PAMAP2, which has 2049 segmented samples per subject on average. However, this dataset comes to be challenging as it contains 10 classes of activities which is higher than the Opportunity dataset with 6 classes. It might adversely affect the performance due to the more probable imbalanced distribution of the classes. Though, the results in Table \ref{table_results_pamap2}, demonstrate that the proposed micro-mini batch learning technique overcame this challenge. The proposed model dominated other states of the art methods in more than 90\% of the experiments and improved W-F1 measure up to 13\%.

The results on Fig. \ref{lissi_resultstraining_plot} presents a competitive performance of the proposed model over LISSI dataset. It dominated the STL and GFK model in all the experiments and came up close to the supervised learning performance. It can be inferred from the results that both adversarial models overall performed well on this imbalanced small size dataset in terms of W-F1 measure.

\begin{table*}[t]
\caption{The main classification metrics of a sample classification task on Opportunity, LISSI, and PAMAP2 datasets.}
\label{classification_reports}
\centering
\resizebox{\textwidth}{!}{
\begin{tabular}{lllr|clccrr|lllr}
\multicolumn{4}{c}{Opportunity Dataset}                    & \multicolumn{1}{c}{} & \multicolumn{4}{c}{LISSI Dataset}                          & \multicolumn{1}{c}{} & \multicolumn{4}{c}{PAMAP2 Dataset}                   \\ \hline \hline  \\[-1.5ex]
\multicolumn{4}{c}{Subject 4 $\rightarrow$ Subject 2}      & \multicolumn{1}{c}{} & \multicolumn{4}{c}{Subject 1  $\rightarrow$  Subject 5}      & \multicolumn{1}{c}{} & \multicolumn{4}{c}{Subject 5  $\rightarrow$  Subject 6}                 \\ \hline \\[-1.5ex]
              class   & precision & recall & support &                      & class        & precision & recall & support &                      &               class   & precision & recall & support \\ \hline
                      &           &        &         &                      &              &           &        &         &                      &                       &           &        &         \\
Null        & \textbf{0.98} & 0.56  & 292     &       & Kneeling   & 0.63      & 0.83      & 53      &    & Ironing  & 0.87      & 0.99      & 137     \\
Relaxing    & 0.81 & \textbf{0.84}  & 190    &      & Lying      & 0.78      & 0.78      & 79      &    & Lying    & 0.89      & \textbf{0.99 }       & 85      \\
Coffee time & 0.34      & 0.33    & 225    &   & Relaxing   & \textbf{0.99} & 0.87    & 142     &    & Sitting  & 0.86      & 0.99       & 84      \\
Early morning & 0.61      & 0.84    & 372     &       & Sitting    & 0.74      & 0.47 & 49      &    & Standing & 0.88      & 0.65       & 89      \\
Clean up       & 0.34      & 0.57   & 224     &     & Sit to Stand  & 0.80   & \textbf{0.88}     & 100    &    & Walking  & 0.87      & 0.86        & 93      \\
Sandwich time & 0.94      & 0.71    & 679     &       & Standing   & 0.87      & 0.71     & 76      &    & Running  & 0.82      & 0.94       & 82      \\
         &           &        &          &       & Dance walk & 0.69      & 0.51      & 57      &    & Cycling  & \textbf{0.93}      & 0.88      & 74      \\
         &           &        &          &       & Warm up    & 0.38      & 0.73       & 41      &    & Ascending & 0.59     & 0.82        & 49      \\
         &           &        &          &       &            &           &        &         &    & Descending & 0.80    & 0.49        & 41      \\
         &           &        &          &       &            &           &        &         &    & Cleaning & 0.70      & 0.43        & 76      \\
         &           &        &          &       &            &           &        &         &    &          &           &                  &          \\[-1.5ex] \hline
Accuracy & 0.67      &        & 1982     &       & Accuracy   & 0.76      &        & 597     &    & Accuracy & 0.83      &                & 810     \\
W-Avg    & 0.74      & 0.67   & 1982     &       & W-Avg  & 0.79      & 0.76   & 597     &    & W-Avg    & 0.84      & 0.83    & 810   \\

\end{tabular}
}
\end{table*}

\begin{figure}[t]
    \setlength{\fboxrule}{0pt}
    \framebox{
      \parbox{0.95\columnwidth}{
        \centering
        \includegraphics[width=0.95\columnwidth, height=10cm, keepaspectratio]{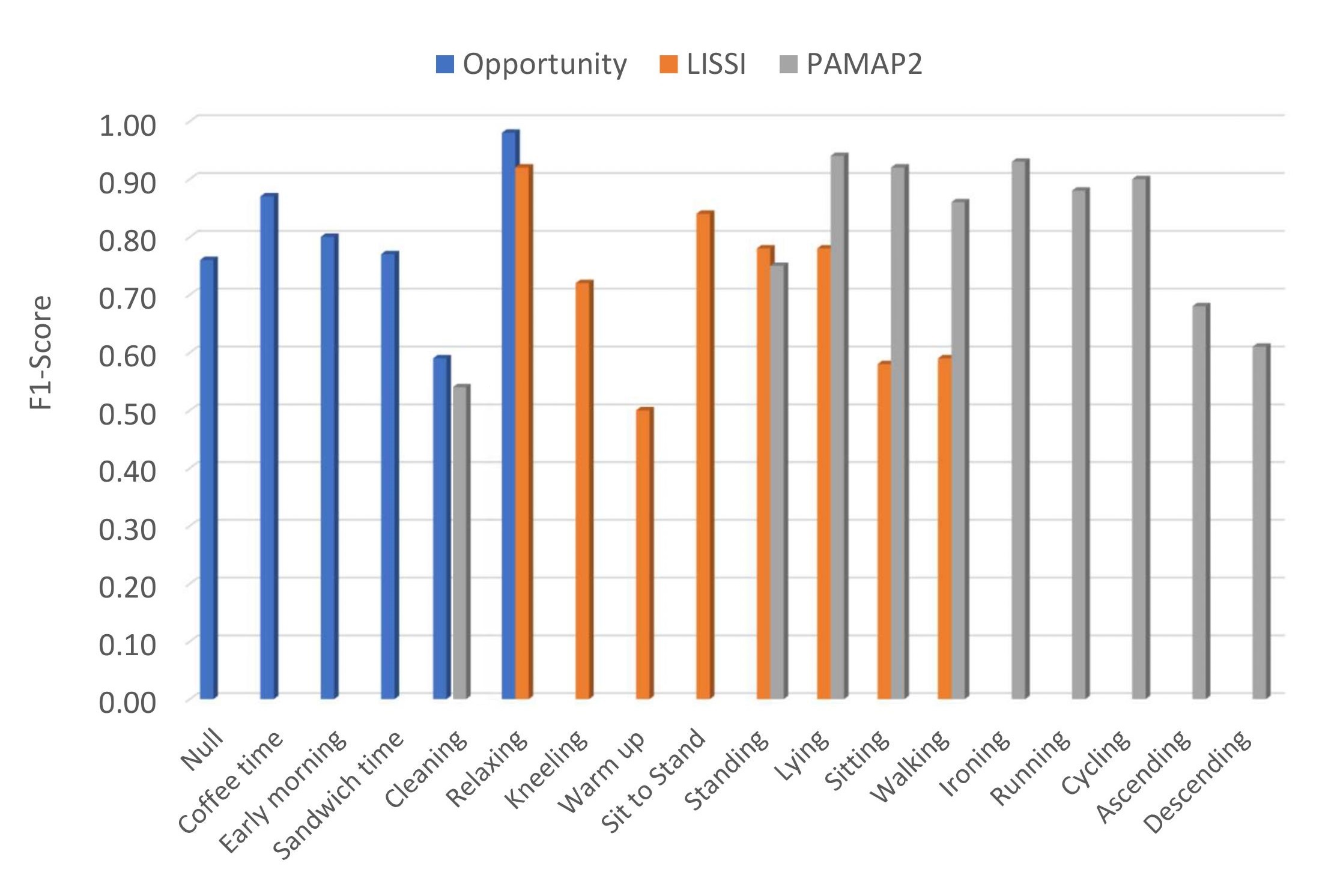}}}
        \caption{Comparison of per class F1-score in Opportunity, PAMAP2, and LISSI datasets.}
        \label{sample_f1_scores}
\end{figure}

\textbf{Performance of the model over a given sample:}
As can be inferred from the Table \ref{classification_reports} and Fig. \ref{sample_f1_scores}, the proposed model performed well in the classification of samples from Subject 2 by leveraging knowledge from subject 4 in Opportunity dataset. The precision values are notably higher than recall and W-F1 measure in \textit{null} and \textit{Sandwich time} classes. Reminding the precision and recall formula as bellow, it can be concluded:
\begin{equation}
\begin{split}
    & Recall = \frac{TP}{TP+FN} , Precision =  \frac{TP}{TP+FP}
   \\& Precision > Recall \rightarrow FN > FP
\end{split}
\end{equation}
which means that the proposed model falsely rejected samples more frequently than it falsely predicted the samples that belong to these two classes. The most likely explanation of this phenomena is the tension of the adversarial models to collapsing, which may push the model to falsely tag samples as if they are drawn from the class that the generator collapsed on. Furthermore, the classifier did not present significant results in recognition of samples from \textit{Coffee time} and \textit{Clean up} classes. It is interesting to note that these classes of activities have been among the most problematic classes, even for a model which had supervised training by the samples of the target subject. This discrepancy could be attributed to their more complicated patterns compared to the rest; the pattern of making coffee may highly vary from a human to another due to the unscripted data acquisition approach \cite{asghari2019online}. Therefore, models require more samples to generalize yet the datasets contain limited number of samples.

The number of samples, overlap between classes,duration of the activities, and level of abstraction are some of the factors associated with the complexity of recognition task.
Supposing slight similarity overlap between classes, the activities contain sharper movements could be straightforward in recognition, since the peaks in acceleration data are more discernible.
According to Table \ref{classification_reports}, the proposed model delivers low precision in recognizing the samples from \textit{Warm up} class of LISSI dataset.
The \textit{Warm up} activity is short-length and contains an extremely rapid sharp movement, which makes the recognition more challenging for the model.

Cleaning activity in both PAMAP2 and Opportunity dataset is amongst the most perplexing activities owing to its high abstraction level.  There exist a large set of sub-activities combination which forms the cleaning activity and no dataset affords full coverage of it. Samples of one activity of the same human subject (domain) may considerably vary based on the innate non-deterministic behavior of human beings; hence, the high generalization capability is required to overcome this intra-domain shift between train and test set of the same domain, as well as the inter-domain shift between source and target domains.
On the other hand, activities such as \textit{Cycling}, \textit{Relaxation}, or \textit{Lying} contain a common inter and intra-domain acceleration pattern. Consequently these activities are more tangible for the model to classify as it can be observed in Table \ref{classification_reports}.

Analysis of Fig. \ref{sample_f1_scores} reveals prevailing recognition performance in terms of F1 measure over PAMAP2 dataset. Nevertheless, a slight phenomena of imbalance learning still can be found over the samples of \textit{Standing}, \textit{Descending} and \textit{Cleaning} class which can be addressed in future work.

\section{Acknowledgement}
\label{Ack}
We thank Prof. Ehsan Nazarfard for the valuable comments he provided during the realisation of the present attempt. We thank also Roghayeh Mojarad, Hazem Abdelkawy, Mohsen Kotobi, Aymen Meziani and Prof. Attal for their implication in the creation of the LISSI dataset during the Medolution project.

\section{Conclusions and Future Work}
\label{future_works}
Recent trends in Generative Adversarial Networks have led to a proliferation of studies that offer adversarial solutions for a variety of applications in Artificial Intelligence mostly Image Processing and Vision.
This study set out to propose a generic adversarial framework for knowledge transfer in the domain of Human Activity Recognition.  The proposed semi-supervised model has been evaluated against three datasets with different challenges to assess its robustness to the scale and imbalance of the data. The findings of our research are quite convincing, and thus the following conclusions can be drawn:

Compared to the SA-GAN adversarial domain adaptation method, our proposed model enhances the final classification performance with an average of  7.5\% for the three datasets, which reinforce the effectiveness of micro-mini-batch training approach.
The proposed model provides striking results on the PAMAP2 benchmark medium-size multi-class dataset. It improved the adversarial domain adaptation performance by applying a micro-mini-batch learning technique on Opportunity large-scale yet highly imbalanced dataset. Interestingly, the proposed model revealed competitive results compared to other states of the art models on the LISSI dataset, which is very challenging in terms of both the number of samples and balance of classes. Our comprehensive assessment was carried out over high-dimensional data of highly abstract activities in all three datasets. Besides, the proposed approach is not HAR-exclusive and it can be potentially utilized to solve other domain adaptation problems.

The results support the effectiveness of the proposed model to address the imbalanced learning challenges. Further studies with more focus on the lack of samples problem will be undertaken. In future investigations, it might be possible to use multiple sources of knowledge or a combination of transferred models from different source domains, and the source/model selection policies. In addition, the integration of the ontology-based reasoning to the present approach could be a mean of improvement of the classification results obtained by Machine Learning.

\bibliographystyle{elsarticle-num}
\bibliography{mybibfile}

\begin{thebibliography}{10}
\expandafter\ifx\csname url\endcsname\relax
  \def\url#1{\texttt{#1}}\fi
\expandafter\ifx\csname urlprefix\endcsname\relax\def\urlprefix{URL }\fi
\expandafter\ifx\csname href\endcsname\relax
  \def\href#1#2{#2} \def\path#1{#1}\fi

\bibitem{FISCHER2020103285}
G.~S. Fischer, R.~da~Rosa~Righi, G.~de~Oliveira~Ramos, C.~A. da~Costa, J.~J.
  Rodrigues,
  \href{http://www.sciencedirect.com/science/article/pii/S0952197619302465}{Elhealth:
  Using internet of things and data prediction for elastic management of human
  resources in smart hospitals}, Engineering Applications of Artificial
  Intelligence 87 (2020) 103285 (2020).
\newblock \href
  {https://doi.org/https://doi.org/10.1016/j.engappai.2019.103285}
  {\path{doi:https://doi.org/10.1016/j.engappai.2019.103285}}.
\newline\urlprefix\url{http://www.sciencedirect.com/science/article/pii/S0952197619302465}

\bibitem{wang2018deep}
J.~Wang, Y.~Chen, S.~Hao, X.~Peng, L.~Hu, Deep learning for sensor-based
  activity recognition: A survey, Pattern Recognition Letters 119 (2019) 3--11
  (2019).

\bibitem{NUNEZ201880}
J.~C. Núñez, R.~Cabido, J.~J. Pantrigo, A.~S. Montemayor, J.~F. Vélez,
  \href{http://www.sciencedirect.com/science/article/pii/S0031320317304405}{Convolutional
  neural networks and long short-term memory for skeleton-based human activity
  and hand gesture recognition}, Pattern Recognition 76 (2018) 80 -- 94 (2018).
\newblock \href {https://doi.org/https://doi.org/10.1016/j.patcog.2017.10.033}
  {\path{doi:https://doi.org/10.1016/j.patcog.2017.10.033}}.
\newline\urlprefix\url{http://www.sciencedirect.com/science/article/pii/S0031320317304405}

\bibitem{chibani2013}
A.~Chibani, Y.~Amirat, S.~Mohammed, E.~Matson, N.~Hagita, M.~Barreto,
  Ubiquitous robotics: Recent challenges and future trends, Robotics and
  Autonomous Systems 61 (2013) 1162--1172 (11 2013).
\newblock \href {https://doi.org/10.1016/j.robot.2013.04.003}
  {\path{doi:10.1016/j.robot.2013.04.003}}.

\bibitem{bengio2013deep}
Y.~Bengio, Deep learning of representations: Looking forward, in: International
  Conference on Statistical Language and Speech Processing, Springer, 2013, pp.
  1--37 (2013).

\bibitem{pan2010survey}
S.~J. Pan, Q.~Yang, A survey on transfer learning, IEEE Transactions on
  knowledge and data engineering 22~(10) (2010) 1345--1359 (2010).

\bibitem{Kouw2019}
W.~M. {Kouw}, M.~{Loog}, A review of domain adaptation without target labels,
  IEEE Transactions on Pattern Analysis and Machine Intelligence (2019) 1--1
  (2019).
\newblock \href {https://doi.org/10.1109/TPAMI.2019.2945942}
  {\path{doi:10.1109/TPAMI.2019.2945942}}.

\bibitem{khan2018scaling}
M.~A. A.~H. Khan, N.~Roy, A.~Misra, Scaling human activity recognition via deep
  learning-based domain adaptation, in: 2018 IEEE International Conference on
  Pervasive Computing and Communications (PerCom), IEEE, 2018, pp. 1--9 (2018).

\bibitem{yosinski2014transferable}
J.~Yosinski, J.~Clune, Y.~Bengio, H.~Lipson, How transferable are features in
  deep neural networks?, in: Advances in neural information processing systems,
  2014, pp. 3320--3328 (2014).

\bibitem{Soleimani2019Cross}
E.~Soleimani, E.~Nazerfard, Cross-subject transfer learning in human activity
  recognition systems using generative adversarial networks, arXiv preprint
  arXiv:1903.12489 (2019).

\bibitem{chavarriaga2013opportunity}
R.~Chavarriaga, H.~Sagha, A.~Calatroni, S.~T. Digumarti, G.~Tr{\"o}ster,
  J.~d.~R. Mill{\'a}n, D.~Roggen, The opportunity challenge: A benchmark
  database for on-body sensor-based activity recognition, Pattern Recognition
  Letters 34~(15) (2013) 2033--2042 (2013).

\bibitem{reiss2012introducing}
A.~Reiss, D.~Stricker, Introducing a new benchmarked dataset for activity
  monitoring, in: 2012 16th International Symposium on Wearable Computers,
  IEEE, 2012, pp. 108--109 (2012).

\bibitem{LISSIdataset}
{LISSI Dataset} rehabilitation self exercises dataset,
  \url{http://www.lissi.fr/parkinson-rehabilitation-dataset-2/}, accessed:
  2019-09-30.

\bibitem{wang2018deeptransfer}
J.~Wang, V.~W. Zheng, Y.~Chen, M.~Huang, Deep transfer learning for
  cross-domain activity recognition, in: Proceedings of the 3rd International
  Conference on Crowd Science and Engineering, ACM, 2018, p.~16 (2018).

\bibitem{hossain2017active}
H.~S. Hossain, M.~A. A.~H. Khan, N.~Roy, Active learning enabled activity
  recognition, Pervasive and Mobile Computing 38 (2017) 312--330 (2017).

\bibitem{pan2011domain}
S.~J. Pan, I.~W. Tsang, J.~T. Kwok, Q.~Yang, Domain adaptation via transfer
  component analysis, IEEE Transactions on Neural Networks 22~(2) (2011)
  199--210 (2011).

\bibitem{wang2018stratified}
J.~Wang, Y.~Chen, L.~Hu, X.~Peng, S.~Y. Philip, Stratified transfer learning
  for cross-domain activity recognition, in: 2018 IEEE International Conference
  on Pervasive Computing and Communications (PerCom), IEEE, 2018, pp. 1--10
  (2018).

\bibitem{gong2012geodesic}
B.~Gong, Y.~Shi, F.~Sha, K.~Grauman, Geodesic flow kernel for unsupervised
  domain adaptation, in: 2012 IEEE Conference on Computer Vision and Pattern
  Recognition, IEEE, 2012, pp. 2066--2073 (2012).

\bibitem{wang2017balanced}
J.~Wang, Y.~Chen, S.~Hao, W.~Feng, Z.~Shen, Balanced distribution adaptation
  for transfer learning, in: 2017 IEEE International Conference on Data Mining
  (ICDM), IEEE, 2017, pp. 1129--1134 (2017).

\bibitem{khan2017transact}
M.~A. A.~H. Khan, N.~Roy, Transact: Transfer learning enabled activity
  recognition, in: 2017 IEEE International Conference on Pervasive Computing
  and Communications Workshops (PerCom Workshops), IEEE, 2017, pp. 545--550
  (2017).

\bibitem{chen2019cross}
Y.~Chen, J.~Wang, M.~Huang, H.~Yu, Cross-position activity recognition with
  stratified transfer learning, Pervasive and Mobile Computing 57 (2019) 1--13
  (2019).

\bibitem{goodfellow2014generative}
I.~Goodfellow, J.~Pouget-Abadie, M.~Mirza, B.~Xu, D.~Warde-Farley, S.~Ozair,
  A.~Courville, Y.~Bengio, Generative adversarial nets, in: Advances in neural
  information processing systems, 2014, pp. 2672--2680 (2014).

\bibitem{nie2017medical}
D.~Nie, R.~Trullo, J.~Lian, C.~Petitjean, S.~Ruan, Q.~Wang, D.~Shen, Medical
  image synthesis with context-aware generative adversarial networks, in:
  International Conference on Medical Image Computing and Computer-Assisted
  Intervention, Springer, 2017, pp. 417--425 (2017).

\bibitem{schlegl2017unsupervised}
T.~Schlegl, P.~Seeb{\"o}ck, S.~M. Waldstein, U.~Schmidt-Erfurth, G.~Langs,
  Unsupervised anomaly detection with generative adversarial networks to guide
  marker discovery, in: International Conference on Information Processing in
  Medical Imaging, Springer, 2017, pp. 146--157 (2017).

\bibitem{reed2016generative}
S.~Reed, Z.~Akata, X.~Yan, L.~Logeswaran, B.~Schiele, H.~Lee, Generative
  adversarial text to image synthesis, arXiv preprint arXiv:1605.05396 (2016).

\bibitem{zhang2017stackgan}
H.~Zhang, T.~Xu, H.~Li, S.~Zhang, X.~Wang, X.~Huang, D.~N. Metaxas, Stackgan:
  Text to photo-realistic image synthesis with stacked generative adversarial
  networks, in: Proceedings of the IEEE International Conference on Computer
  Vision, 2017, pp. 5907--5915 (2017).

\bibitem{bidgoli2019deepcloud}
A.~Bidgoli, P.~Veloso, Deepcloud. the application of a data-driven, generative
  model in design, arXiv preprint arXiv:1904.01083 (2019).

\bibitem{ueda2018data}
T.~Ueda, M.~Seo, I.~Nishikawa, Data correction by a generative model with an
  encoder and its application to structure design, in: International Conference
  on Artificial Neural Networks, Springer, 2018, pp. 403--413 (2018).

\bibitem{SenseGen2017}
M.~Alzantot, S.~Chakraborty, M.~B. Srivastava, Sensegen: A deep learning
  architecture for synthetic sensor data generation, 2017 IEEE International
  Conference on Pervasive Computing and Communications Workshops (PerCom
  Workshops) (2017) 188--193 (2017).

\bibitem{saeedi2018personalized}
R.~Saeedi, K.~Sasani, S.~Norgaard, A.~H. Gebremedhin, Personalized human
  activity recognition using wearables: A manifold learning-based knowledge
  transfer, in: 2018 40th Annual International Conference of the IEEE
  Engineering in Medicine and Biology Society (EMBC), IEEE, 2018, pp.
  1193--1196 (2018).
\newblock \href {https://doi.org/10.1109/EMBC.2018.8512533}
  {\path{doi:10.1109/EMBC.2018.8512533}}.

\bibitem{wang2018sensorygans}
J.~Wang, Y.~Chen, Y.~Gu, Y.~Xiao, H.~Pan, Sensorygans: An effective generative
  adversarial framework for sensor-based human activity recognition, in: 2018
  International Joint Conference on Neural Networks (IJCNN), IEEE, 2018, pp.
  1--8 (2018).

\bibitem{tzeng2017adversarial}
E.~Tzeng, J.~Hoffman, K.~Saenko, T.~Darrell, Adversarial discriminative domain
  adaptation, in: Proceedings of the IEEE Conference on Computer Vision and
  Pattern Recognition, 2017, pp. 7167--7176 (2017).
\newblock \href {https://doi.org/10.1109/CVPR.2017.316}
  {\path{doi:10.1109/CVPR.2017.316}}.

\bibitem{luo2017label}
Z.~Luo, Y.~Zou, J.~Hoffman, L.~F. Fei-Fei, Label efficient learning of
  transferable representations acrosss domains and tasks, in: Advances in
  Neural Information Processing Systems, 2017, pp. 165--177 (2017).

\bibitem{bousmalis2017unsupervised}
K.~Bousmalis, N.~Silberman, D.~Dohan, D.~Erhan, D.~Krishnan, Unsupervised
  pixel-level domain adaptation with generative adversarial networks, in:
  Proceedings of the IEEE conference on computer vision and pattern
  recognition, 2017, pp. 3722--3731 (2017).

\bibitem{wang2019deep}
J.~Wang, Y.~Chen, S.~Hao, X.~Peng, L.~Hu, Deep learning for sensor-based
  activity recognition: A survey, Pattern Recognition Letters 119 (2019) 3--11
  (2019).

\bibitem{jiang2008literature}
J.~Jiang, A literature survey on domain adaptation of statistical classifiers,
  URL: http://sifaka. cs. uiuc. edu/jiang4/domainadaptation/survey 3 (2008)
  1--12 (2008).

\bibitem{srivastava2017veegan}
A.~Srivastava, L.~Valkov, C.~Russell, M.~U. Gutmann, C.~Sutton, Veegan:
  Reducing mode collapse in gans using implicit variational learning, in:
  Advances in Neural Information Processing Systems, 2017, pp. 3308--3318
  (2017).

\bibitem{salimans2016improved}
T.~Salimans, I.~Goodfellow, W.~Zaremba, V.~Cheung, A.~Radford, X.~Chen,
  Improved techniques for training gans, in: Advances in neural information
  processing systems, 2016, pp. 2234--2242 (2016).

\bibitem{arjovsky_towards_2017}
M.~Arjovsky, L.~Bottou, Towards principled methods for training generative
  adversarial networks, arXiv preprint arXiv:1701.04862 (2017).

\bibitem{asghari2019online}
P.~Asghari, E.~Soleimani, E.~Nazerfard, Online human activity recognition
  employing hierarchical hidden markov models, Journal of Ambient Intelligence
  and Humanized Computing (2019).
\newblock \href {https://doi.org/10.1007/s12652-019-01380-5}
  {\path{doi:10.1007/s12652-019-01380-5}}.

\end{thebibliography}
\balance

\end{document}